# An Interpretable Benchmark for Clickbait Detection and Tactic Attribution

Lihi Nofar
*School of Computer Science,
Faculty of Sciences*
Holon Institute of Technology

Tomer Portal
*School of Computer Science,
Faculty of Sciences*
Holon Institute of Technology

Aviv Elbaz
*School of Computer Science,
Faculty of Sciences*
Holon Institute of Technology

Alexander Apartsin
*School of Computer Science,
Faculty of Sciences*
Holon Institute of Technology

Yehudit Aperstein
*Intelligent Systems,
Afeka Academic College of Engineering*
Tel Aviv Israel

**The proliferation of clickbait headlines poses significant challenges to the credibility of information and user trust in digital media. While recent advances in machine learning have improved the detection of manipulative content, the lack of explainability limits their practical adoption. This paper presents a model for explainable clickbait detection that not only identifies clickbait titles but also attributes them to specific linguistic manipulation strategies. We introduce a synthetic dataset generated by systematically augmenting real news headlines using a predefined catalogue of clickbait strategies. This dataset enables controlled experimentation and detailed analysis of model behaviour. We present a two-stage framework for automatic clickbait analysis comprising detection and tactic attribution. In the first stage, we compare a fine-tuned BERT classifier with large language models (LLMs), specifically GPT-4.0 and Gemini 2.4 Flash, under both zero-shot prompting and few-shot prompting enriched with illustrative clickbait headlines and their associated persuasive tactics. In the second stage, a dedicated BERT-based classifier predicts the specific clickbait strategies present in each headline. This work advances the development of transparent and trustworthy AI systems for combating manipulative media content. We share the dataset with the research community at https://github.com/LLM-HITCS25S/ClickbaitTacticsDetection**

## I. INTRODUCTION

The widespread use of clickbait headlines in digital media has become a pervasive challenge, undermining the credibility of information and exploiting user attention through manipulative linguistic techniques. While automated systems for detecting clickbait have improved in recent years, their focus has remained mainly on binary classification, simply labelling content as clickbait or not. However, effective mitigation of such content requires going beyond detection to understanding *how* and *why* certain headlines manipulate readers. Specifically, it is crucial to evaluate whether current AI models can accurately recognize and distinguish the diverse linguistic styles and persuasive strategies commonly employed in clickbait.

In this study, we systematically benchmark the capabilities of modern AI models, both fine-tuned classifiers and large language models (LLMs), to detect clickbait content and *distinguish among the underlying rhetorical strategies employed in its construction*. To support this analysis, we generate a synthetic dataset by augmenting real news headlines with controlled applications of predefined clickbait strategies, such as exaggeration, curiosity gaps, emotional triggers, and misleading claims. This dataset enables a rigorous evaluation of models' abilities to perform not only binary detection but also detailed *strategy attribution*.

In this work, we present a two-stage framework for analyzing clickbait headlines. The first stage focuses on clickbait detection, while the second stage performs tactic attribution, identifying the specific persuasive strategies used in each headline. This separation allows us not only to classify a headline as clickbait or non-clickbait, but also to explain *why* a given headline is manipulative.

For the detection stage, we evaluate a range of models and prompting strategies. We include a fine-tuned BERT classifier, as well as state-of-the-art large language models (LLMs) such as GPT-4.0 and Gemini 2.4 Flash. These LLMs are tested in both zero-shot settings, where the model receives only task instructions, and few-shot settings, where the prompt is enriched with explicit examples of clickbait headlines and their associated tactics. This design allows us to examine how providing tactic-aware exemplars influences detection performance.

The attribution stage employs a dedicated BERT-based classifier trained to predict the specific clickbait tactics present in each headline. By separating detection from attribution, we obtain interpretable outputs that highlight the particular persuasive strategies, such as curiosity-gap framing or sensational adjectives, that drive engagement.

By benchmarking these approaches, we provide valuable insights into the current state of explainable AI for clickbait detection and the models' abilities to perform fine-grained linguistic attribution. This work contributes toward the development of transparent and interpretable AI systems capable of identifying not only manipulative content but also the techniques that drive it, supporting efforts to promote media integrity and user trust.

## II. LITERATURE REVIEW

Clickbait refers to sensational or misleading headlines crafted to provoke curiosity and entice readers to click. These headlines often promise information that the linked content fails to deliver, degrading user trust. In recent years, research on automatically detecting clickbait has grown, alongside a rising emphasis on explainability, i.e., providing human-interpretable reasons for why a headline is classified as clickbait. Explainable detection is crucial for enhancing user trust in AI systems and enabling users to understand the linguistic strategies behind clickbait.

Research on clickbait has benefited from several datasets of headlines labelled as clickbait or non-clickbait. A foundational resource is the Clickbait Challenge 2017 dataset, compiled by the Webis group, which contains thousands of social media posts with associated clickbait scores. Since then, new datasets have extended to other languages and domains. For example, BanglaBait (Mahtab et al., 2023) introduced a Bangla-language headline corpus with clickbait labels, and RoCliCo (Broscoteanu & Ionescu, 2023) provided a Romanian clickbait news headline corpus/ Researchers have also compiled clickbait datasets from online news in Turkish (Akgün & İnkaya, 2024) and Chinese (Liu et al., 2022), reflecting a trend toward multilingual clickbait detection. Many datasets focus on short text, such as headlines or social media post text, sometimes accompanied by additional context, like article snippets or images (for multi-modal clickbait).

Clickbait is related to misinformation tasks, so we briefly note some key datasets in this context. The FakeNewsNet repository (Shu et al., 2018) aggregates news articles labelled as fake or real, along with their social context. It has been utilized in explainable fake news studies, such as dEFEND (Shu et al., 2019). Short-form misinformation is addressed by datasets such as LIAR, which labels political statements with truth values, and the SemEval-2019 Hyperpartisan News dataset, which categorizes news articles based on extreme bias. Another relevant task is propaganda detection, where datasets from 2019 to 2021 label spans of text that employ propaganda techniques. These serve as a fine-grained benchmark for explainable classification, as the goal is to both detect propaganda and highlight the specific words or phrases (e.g., slogans, name-calling) responsible. In summary, a variety of datasets exist, enabling supervised learning for clickbait and fake news detection across languages and providing testbeds for the development of explanation methods. However, data scarcity remains a significant issue for low-resource languages and complex multimodal scenarios, motivating the need for data augmentation and synthetic data generation, as discussed later.

Early approaches (pre-2019) relied heavily on hand-crafted features to capture clickbait cues, such as unusual punctuation (e.g., excessive "!" or "?"), the presence of forward-reference words like "this" or "someone" (creating curiosity gaps), and the use of superlatives or suspenseful language. Over the last five years, several studies have continued to explore feature-driven methods, partly to facilitate cross-language generalization. For example, Coste and Bufnea (2021) proposed a language-independent strategy for clickbait detection, utilizing features such as punctuation patterns and stop-word ratios, which can be applied across languages. Traditional classifiers, such as logistic regression, SVM, and random forests, built on lexical and semantic features, have been evaluated on clickbait data as baselines for comparison. Yadav and Bansal (2023) conducted a comparative study of machine learning methods for clickbait, confirming that ensembles of lexical features can achieve reasonable accuracy, though deep learning approaches generally outperform them. Overall, feature-based models help illuminate which linguistic attributes are most predictive, such as specific parts-of-speech (POS) tags or phrase patterns. However, they require expert knowledge for feature engineering and may struggle with the nuanced or context-dependent nature of modern clickbait.

Modern clickbait detectors primarily utilize deep learning to learn features from data automatically. Recurrent neural networks (RNNs) and convolutional neural networks (CNNs) were among the first such models applied to clickbait headlines, demonstrating improved performance by capturing word sequence patterns that simple features missed. For example, Anand et al. (2017) used an LSTM-based model, which showed gains by learning temporal word dependencies. In contrast, CNN models (e.g., Agrawal, 2016) captured keyword presence through n-gram filters. Later works incorporated attention mechanisms to focus on the most salient words in a headline. Mishra et al. (2020) introduced an attention-based network that highlighted important trigger words (e.g., "will not believe" or "shocking") to improve detection accuracy. Graph Neural Networks have also been explored; Liu et al. (2022) designed a Graph Attention Network that integrates semantic relations and syntactic structure (dependency links) of words.

Their WeChat clickbait detector combined a BERT-based text encoder with features such as part-of-speech tags to capture both meaning and form, significantly enhancing performance on a dataset of Chinese social media headlines (Liu et al., 2022). In cases where social context or multiple modalities are available, deep models have been extended to use them. Some approaches consider user behavior signals: for instance, metadata such as the number of clicks, comments, or shares can hint at clickbaitness. However, relying on propagation patterns means detecting clickbait only after it has spread, which is often too late for mitigation. Thus, content-based methods are preferred for early detection. Multimodal models combine textual analysis with image analysis when posts include an enticing thumbnail image. Researchers have noted that inconsistency between a headline and its accompanying image (or article content) is a strong indicator of clickbait. For example, a post showing an unrelated sensational image alongside a headline can be flagged. Yu et al. (2024) propose a causal inference-based multimodal model that disentangles intrinsic "bait" content from irrelevant content introduced to fool detectors. By identifying latent factors representing the true deceptive intention, their model became more robust to "disguised" clickbait, where creators insert innocuous text or images to evade detection. Overall, deep learning has enabled highly accurate clickbait classifiers by learning complex feature combinations; yet these models are often opaque, which drives the need for explainability techniques.

The introduction of large pre-trained language models (PLMs), such as BERT and RoBERTa, led to further improvements in clickbait detection between 2019 and 2022. Fine-tuning such models on clickbait datasets leverages their vast linguistic knowledge. Indurthi et al. (2020) demonstrated that a fine-tuned BERT can predict not only binary clickbait versus non-clickbait, but also a clickbait score (a graded measure of clickbait strength) with high accuracy. Similarly, Yi et al. (2022) proposed a contrastive variational modelling approach built on PLMs, which treats clickbait detection as learning distinct representations for clickbait versus normal headlines. By using contrastive learning, their IJCAI 2022 model improved the discrimination of subtle cues. Fine-tuned transformers have become the standard baselines, often achieving F1-scores above 0.85 on English benchmarks and outperforming earlier RNN and CNN models. For low-resource languages, multilingual PLMs (e.g., mBERT, XLM) have been fine-tuned when limited native data is available. For example, mBERT was used in some Turkish and Bangla clickbait studies to leverage cross-lingual transfer.

One challenge noted with PLM fine-tuning is the potential gap between the pre-training objective and the clickbait detection task, which can limit performance if sufficient task-specific data is not present. To mitigate data scarcity, researchers have explored data augmentation techniques to enhance the availability of data. Yoon et al. (2019) injected augmented samples (paraphrased or noised headlines) to expand training data, while López-Sánchez et al. (2018) experimented with domain adaptation to transfer knowledge from news headlines to social media posts. These strategies can help a fine-tuned model generalize better, though care must be taken to avoid introducing spurious patterns in augmented data.

By 2023, the focus shifted to large language models (LLMs), such as GPT-3 and ChatGPT, to determine whether they can detect clickbait without task-specific training. Wang et al. (2023) evaluated GPT-3.5 (via ChatGPT) in zero-shot and few-shot settings on multiple English and Chinese datasets. They found that zero-shot LLM prompts (e.g., asking ChatGPT, "Is this headline clickbait? Yes, or no?") achieve reasonable accuracy but do not yet match the accuracy of fine-tuned specialized models. In particular, contrary to human intuition that a headline alone often suffices to judge clickbait, GPT-based models struggled to do so reliably with just a prompt. Few-shot prompting, which involves providing a few labelled examples in the prompt, improved performance but still fell short of the state-of-the-art supervised methods. One insight from these experiments is that LLMs have strong multilingual abilities – ChatGPT's performance was relatively consistent across English and Chinese clickbait, indicating that it can generalize its understanding of enticing language to some degree.

Another line of work that integrates prompt learning for clickbait is Qin et al. (2022), as referenced in a Chinese social media study, which introduced tailored prompt templates to stimulate the knowledge of PLMs for Weibo clickbait detection. These prompt-based fine-tuning approaches, which fall between zero-shot and full fine-tuning, have shown promise in reducing the amount of training data required. In summary, off-the-shelf large language models (LLMs) today provide a convenient baseline for clickbait detection, which is suitable for rapid deployment in new domains; however, specialized training or adaptation is still necessary to achieve optimal performance. As LLMs continue to improve, future systems may combine their broad knowledge with fine-tuning tailored to clickbait for both high accuracy and explainability.

A central theme in recent literature is explainable AI (XAI) for content classification. Simply detecting clickbait or fake news is often not enough – platforms and end-users want to know why a piece of content was flagged. This has led to various approaches that provide attributions, highlights, or natural-language explanations alongside predictions. Feature Attribution (Post-hoc Explanations): A straightforward way to explain a clickbait classification is to identify which features or words most influenced the model's decision. General XAI tools, such as LIME and SHAP, have been applied in this context. For instance, Akgün and İnkaya (2024) used SHAP values to explain a machine learning classifier for Turkish clickbait. By inputting a headline into the model and computing SHAP, they could rank the words or n-grams that contributed most to a "clickbait" prediction (e.g., words like "shocking" or "you will not believe" might get high positive attributions). Such post-hoc explanations help uncover common patterns the model relies on. In their study, the SHAP analysis confirmed that specific sensational and curiosity-inducing terms in Turkish were key factors in determining the results.

Similarly, one could apply LIME (Ribeiro et al., 2016) to generate an explanation by perturbing parts of the headline and observing the impact on the prediction. While these methods do not alter the model's functionality, they provide a transparent layer on top of an otherwise opaque model. A limitation, however, is that the reliability of these explanations depends on the stability of the model's behavior under small changes; nonetheless, they are popular for quick insights. Attention and Rationale Extraction: In neural network models that use attention mechanisms or similar architectures, the model itself can produce internal attributions. Attention weights can highlight which words in a headline were most attended to when predicting clickbait. For example, an attention-based LSTM might implicitly show that words like "Top 10" or "Guess what" in the title drove the classification. Caution is needed since attention weights are not always faithful explanations, but they often align with intuitive features. Some research goes further by building rationale extraction directly into the model. Shu et al. (2019) in dEFEND pioneered this approach for fake news, employing dual attention (co-attention) over news content and user comments to identify the top-k snippets (sentences and comments) that serve as explanations for why a news article is predicted to be fake. The output is not just a label but also a set of highlighted sentences (e.g., a dubious claim in the article and a user comment debunking it) that explain the decision. Notably, they quantitatively evaluated explanation quality by checking how well the extracted sentences matched human-identified check-worthy sentences, using metrics like NDCG and Precision.

This concept of extractive explanations has also influenced clickbait detection. While headlines are short, some approaches utilize the linked article or social media discussion as additional context and identify inconsistent or revealing pieces of text to support the clickbait prediction. Another example is Chien et al. (2022), who developed XFlag, an explainable fake news model that provides reasons for a verdict on social media posts. Although details of XFlag are beyond our scope, it integrates stance detection to flag conflicting information between a post and known facts, thereby explaining (Chien et al., 2022). In general, multi-modal explanations are also explored: a system could highlight an image (or portion of it) that does not match the headline, explaining that "the image is unrelated to the claim." Such capabilities remain an active area of research.

By 2023–2024, an emerging trend is to leverage large language models themselves to generate explanations for decisions. Huang et al. (2024) (FakeGPT) treated ChatGPT not only as a detector but also as an explainer for fake news. They prompted ChatGPT to justify its classification of news as fake or real, and extracted common rationale patterns from these explanations. Interestingly, by analyzing ChatGPT's free-text explanations for multiple instances, they identified nine key features characteristic of fake news, including references to a lack of credible sources and the presence of clickbait language. This demonstrates a novel use of LLMs: the model's explanation can be parsed to identify which factors (like "the headline makes an exaggerated claim" or "the content contradicts known facts") are driving the decision. Those factors can then be evaluated across datasets to see how frequently they occur in fake vs real news. This kind of approach bridges human-understandable concepts and model decisions. Another work used a "reason-aware" prompting technique, where the LLM is asked to consider specific potential reasons before answering, which improved detection consistency.

For clickbait, one could envision ChatGPT explaining, "This headline is likely clickbait because it withholds key information (who, what, or where) to invoke curiosity." Such explanations, if accurate, are beneficial for end-users and content moderators. The downside is that LLM-generated explanations are not guaranteed to reflect the model's internal reasoning accurately; they can sound plausible but be incorrect. Hence, there is interest in evaluating explanation quality – e.g., measuring whether the provided reason is truthful and aligns with known evidence, or conducting user studies to determine whether the explanation increases understanding and trust. Evaluation of Attribution: Regardless of the method, evaluating explainability is a challenging task. Some objective metrics have been used, like fidelity (does removing the highlighted features significantly reduce the model's confidence?) or precision@k for rationale retrieval (as in dEFEND, checking if the top-k highlighted sentences indeed contain misinformation cues).

A crucial aspect of explainable clickbait detection is understanding the persuasive linguistic strategies that differentiate clickbait from normal headlines. Researchers in communications and computational linguistics have identified a variety of recurring patterns in clickbait headlines. Clickbait often withholds key details, using phrases like "You will not believe…" or "This [person/thing] …" to spark curiosity. Blom and Hansen (2015) termed this forward-reference, as the headline refers to something (e.g., "this actress") without naming it, forcing the reader to click for details. Recent analyses confirm that unanswered questions in headlines and deictic references (like "here" or "this") are standard clickbait devices. For instance, headlines posed as open questions ("Can you guess what happened next?") or listicles ("10 Secrets to...") inherently promise that the complete information is inside, thus baiting the click.

Many clickbait headlines use exaggeration, hyperbole, or sensational language to create excitement or alarm. Words like "insane", "shocking", "heartbreaking", etc., or excessive punctuation ("!!!") are signals. A study by Flórez-Vivar and Zaharía (2022) categorized exaggeration tactics, including neologisms, intensifiers (e.g., "very best", "absolutely stunning"), and the use of all-caps or exclamation marks. Such embellishments aim to make the content seem more extraordinary than it is. Recent psychological studies (e.g., Scott, 2021) note that these emotional triggers exploit the reader's instinct for novel or extreme information (related to relevance theory and information gap theory in communication)

Clickbait frequently addresses the reader directly or uses a conversational tone, unlike traditional news headlines. This can include second-person pronouns ("you"), imperatives ("watch this now"), or colloquial expressions. Using personal pronouns and even slang can create a sense of intimacy or urgency. For example, "This hack will change your life" directly appeals to you, the reader. Flórez-Vivar & Zaharía (2022) list personal deixis and imperatives as a key category, along with typographic choices like adding parentheticals or ellipses to create a dramatic pause. Such techniques break the formal tone of standard journalism, signaling that the headline's goal is engagement over information.

Legitimate headlines often provide specific information, such as names, dates, and facts, whereas clickbait tends to lean toward ambiguity to create intrigue. For example, a non-clickbait headline might be, "Mayor John Doe Resigns Amid Scandal," while a clickbait version would be "You will Never Guess Who Just Resigned in Disgrace." The latter omits the key detail, the name, to pique curiosity. Studies have also noted that clickbait may misuse credibility markers (like "Study shows..." without linking to a study) or include misleading frames. In some cases, clickbait borders on false or deceptive content – indeed, a subset of clickbait headlines is essentially fake news with a sensational wrapper. Pasternack (2019) coined the term "deceptive clickbait" for headlines that not only entice but also outright mislead about the content. These require detection techniques that overlap with those used for fake news verification to verify whether claims align with reality.

Related to clickbait, misinformation posts often employ persuasion strategies. For instance, propaganda pieces might use loaded language or anecdotal leads to capture readers' attention. A 2023 study on misinformation in social media (Chen et al., 2023) found that posts with misleading intent often employ similar tactics to clickbait, including emotional appeal, storytelling (utilizing anecdotes to evoke empathy), and logical fallacies, to sway public opinion. Identifying these strategies can help explain why a piece of content was flagged, for example, "the text uses a personal story laced with exaggerated claims, indicative of a persuasive misinformation tactic."

Recent literature surveys confirm the importance of these linguistic features. Jácobo-Morales & Marino-Jiménez (2024) compiled 165 studies and identified 11 key properties of clickbait, including provoking curiosity, employing a sensational or informal style, being misleading or of low quality, and prioritizing click maximization. Likewise, Lu & Shen (2023) identified humor, logical puzzles, storytelling, and other strategies as common elements in clickbait, particularly in video titles. The implication for detection models is that incorporating knowledge of these patterns can enhance the model's explainability. Some detection systems explicitly check for the presence of such phrases or structures as features (e.g., a binary feature for "does the headline pose a question?"). When an explainable model flags a headline, it might output which pattern was detected – for example, "classified as clickbait due to list-based structure and curiosity-provoking phrasing." This not only helps end-users but also assists content creators in avoiding unintentional clickbait styles if desired.

Addressing data limitations has been a recurring theme in recent years, particularly for training models that can be effectively explained and generalized. Researchers have turned to synthetic data generation to supplement real datasets. One approach has been using LLMs to generate additional examples of fake news or clickbait. Huang et al. (2024) demonstrated this by prompting ChatGPT to produce fake news articles in various styles.

They demonstrated, through human evaluation, that the samples generated were of high quality and diversity, effectively augmenting the training data for detectors. Similarly, for clickbait, one can generate artificial clickbait headlines by applying known transformation rules, such as rephrasing a typical headline as a question or adding a teaser phrase. Such synthetic clickbait can be helpful in scenarios where we have an abundance of non-clickbait samples but a scarcity of clickbait examples. Another sophisticated technique is the use of adversarial methods to generate data. Mahtab et al. (2023) in their BanglaBait work employed a semi-supervised adversarial approach, utilizing a generative model to produce candidate clickbait headlines from unlabeled data and an adversarial setup to refine the classifier gradually. This process generated additional training points in low-resource settings, improving the model without requiring manual annotation.

Broscoteanu and Ionescu (2023) applied contrastive learning to the Romanian clickbait corpus, aiming to make more effective use of the limited data. This approach essentially involves learning from pairs of clickbait versus non-clickbait examples by adjusting representations to pull them closer together or apart as needed. In multimodal misinformation detection, Zeng et al. (2024) addressed the lack of image-text training data by learning from synthetic multimodal misinformation. They note that large, curated real-world fact-checking sets, such as MediaEval (with only ~10,000 items), are dwarfed by synthetic sets, like NewsCLIPings (with over 1 million generated image-caption pairs). By carefully selecting and weighting synthetic samples to mimic the distribution of real data, they enhanced the performance of a 13-billion-parameter model in detecting real fake news, even outperforming GPT-4 Vision on some benchmarks. This highlights that synthetic data, when used wisely, can enhance detection systems.

III. METHODOLOGY

This study employs a structured methodology designed to facilitate both accurate clickbait detection and explainable attribution of the linguistic strategies employed in clickbait creation. The methodology consists of three main stages: (1) sourcing and formalizing clickbait creation tactics, (2) generating a synthetic dataset through controlled headline augmentation, and (3) implementing and evaluating model pipelines for detection and attribution.

We begin by constructing a comprehensive catalogue of clickbait creation strategies using a hybrid approach. An AI-assisted analysis is performed on multiple sources, including prior academic research in media psychology, computational linguistics, journalism studies, and large-scale headline corpora. This process identifies recurring persuasive patterns and linguistic manipulations commonly used to increase engagement.

| Tactic | Description |
|---|---|
| Curiosity Gap | Withholding key information to provoke curiosity. |
| Exaggeration and Hyperbole | Overstating claims to increase perceived importance. |
| Emotional Triggers | Using emotionally charged language to elicit strong reactions. |
| Sensationalism | Framing ordinary events as shocking or extraordinary. |
| Lists and Superlatives | Utilizing list structures or superlatives can help attract attention. |
| Ambiguous References | Using vague entities to create intrigue. |
| Direct Appeals | Addressing the reader directly. |
| Unfinished Narratives | Implies a story but leaves out the conclusion. |
| Unexpected Associations | Juxtaposing unrelated concepts to spark curiosity. |
| Provocative Questions | Using open or rhetorical questions to engage the reader. |

**Table 1**: Common Clickbait Tactics

To create a rich and labelled dataset for model training and evaluation, we adopt a controlled synthetic data generation process based on real-world news headlines. A large corpus of verified, non-clickbait news headlines is sampled from a publicly available news headline dataset [37], ensuring broad coverage of topics and writing styles. For each sampled headline, we generate multiple clickbait variants by applying one or more tactics from the predefined catalogue. We employ the GPT-4o model for this task, using carefully crafted prompts that specify the tactic(s) to apply. For example, a prompt might instruct the model to rewrite a headline using the "Curiosity Gap" and "Exaggeration" tactics.

| Source Item | Tactics Used | Augmented Item |
|---|---|---|
| Haiti PM resigns as transitional council is sworn in | Unfinished Narrative | Power Shifts in Haiti as Transitional Council Sworn In—But the Next Move? |
| Meta AI spending plans cause share price slump | Curiosity Gap | Why Did Meta's Huge AI Bet Send Its Stock Tumbling Overnight? |

**Table 2**: Original and clickbait augmented news title examples

Each generated clickbait headline is annotated with the exact set of tactics used during its creation. This ensures that the resulting dataset supports both binary clickbait detection and multi-label classification for tactic attribution. The final dataset includes the original headlines, their clickbait variants, and corresponding tactic labels, enabling supervised learning for both detection and explainability.

To examine the balance between accuracy, interpretability, and computational cost, we evaluate a set of two-stage pipelines for clickbait detection and tactic attribution. In every configuration, the first stage performs headline detection, and the second stage provides tactic attribution. However, the choice of models and prompting strategies varies to illuminate the trade-offs between specialized training and the use of large pre-trained language models.

In the detection stage, we compare several alternatives. The first employs a fine-tuned BERT classifier, representing a fully supervised baseline. The remaining configurations rely on large language models, specifically GPT-4.0 and Gemini 2.5 Flash, each tested under both zero-shot and few-shot prompting conditions. This setup allows us to assess how these state-of-the-art LLMs perform clickbait detection without task-specific fine-tuning and how tactic-aware few-shot exemplars affect their accuracy. We also examine a hybrid configuration that keeps the BERT detector but introduces few-shot prompting to GPT-4.0 for downstream attribution, measuring the impact of minimal tactic-annotated examples on performance.

The attribution stage is handled by a dedicated BERT-based multi-label classifier whenever attribution is not produced directly by the LLM. This component predicts the specific persuasive strategies, such as curiosity gaps or sensational framing, present in each headline. When few-shot prompting is applied, both GPT-4.0 and Gemini 2.5 Flash receive labeled exemplars that illustrate clickbait titles together with their associated tactics, enabling direct tactic prediction alongside detection.

Across all configurations, including those where GPT-4.0 and Gemini 2.5 Flash serve as detectors with both zero-shot and few-shot prompting, we measure detection accuracy, attribution precision, and the clarity of the generated explanations, with particular attention to the system's ability to distinguish overlapping persuasive techniques.

## IV. RESULTS

The proposed pipelines were evaluated using precision, recall, and F1-score to measure both clickbait detection and tactic attribution. Evaluation was conducted at two levels. Table 3 reports the detection performance of all models. A fine-tuned BERT classifier achieved the highest scores, with a precision of 0.89, a recall of 0.93, and an F1-score of 0.91. Large language models showed lower absolute performance but improved when tactic-aware few-shot prompts were used. GPT-4.0 increased its F1-score from 0.40 in zero-shot mode to 0.43 in few-shot mode, while Gemini 2.5 Flash improved from 0.46 to 0.47 under the same setting.

| Detection Model | Accuracy | Precision | Recall | F1-Score |
|---|---|---|---|---|
| Fine-tuned BERT | **0.91** | **0.89** | **0.93** | **0.91** |
| GPT-4o Zero-Shot | 0.73 | 0.31 | 0.55 | 0.40 |
| GPT-4o Few-Shot | 0.84 | 0.36 | 0.53 | 0.43 |
| Gemini 2.5 Few Shots | 0.81 | 0.40 | 0.55 | 0.46 |
| Gemini 2.5 Zero-shot | 0.83 | 0.38 | 0.60 | 0.47 |

**Table 3**: Clickbait Detection Performance

Tactic attribution was evaluated as a multi-label classification task, reporting overall Precision, Recall, and F1-score to capture the accuracy of identifying the persuasive strategies present in each headline (Table 4).

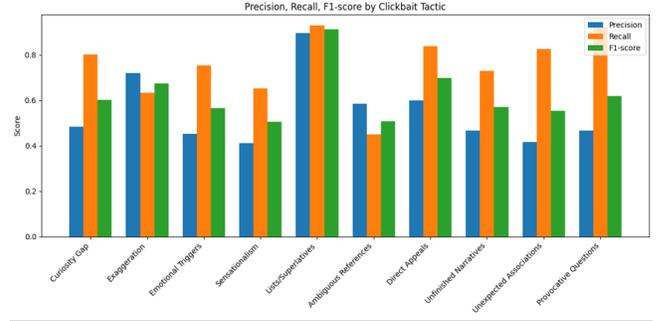

**Table 4**: Attribution Performance

## V. CONCLUSIONS AND FUTURE RESEARCH

This study introduced an explainable framework for clickbait detection that moves beyond simple classification to identify the specific linguistic strategies used to attract readers. We compiled a catalogue of ten common clickbait tactics and created a synthetic dataset through controlled augmentation with GPT-4.0, enabling precise evaluation of both detection and tactic attribution.

Our experiments compared two-stage pipelines built on fine-tuned BERT models and on large language models (GPT-4.0 and Gemini 2.5 Flash), evaluated with both zero-shot and few-shot prompting. The results highlight clear trade-offs: fine-tuned models deliver the highest precision, recall, and F1-scores for detection. In contrast, large language models show strong potential for direct attribution and perform reasonably well without task-specific training, especially when tactic-aware few-shot examples are provided. These findings underscore the balance between accuracy, explainability, and deployment complexity when integrating large pre-trained models.

Future work will broaden the dataset to include multimodal signals such as images and social context to mirror real-world use better. We plan to explore causal inference techniques to separate overlapping persuasive tactics and to enhance attribution precision. Extending the approach to additional languages and low-resource settings remains a priority, along with incorporating human-in-the-loop evaluation to gauge the clarity and usefulness of model explanations for moderators and end-users. Finally, we aim to investigate conversational AI tools that can not only detect and explain clickbait but also offer interactive guidance to help content creators avoid manipulative writing practices.